# Deep Learning-based High-precision Depth Map Estimation from Missing Viewpoints for 360 Degree Digital Holography

**Hakdong Kim[1], Heonyeong Lim[1], Minkyu Jee[2], Yurim Lee[3], Jisoo Jeong[2], Kyudam Choi[2], MinSung Yoon[4, *], and Cheongwon Kim[5, **]**

[1]Department of Digital Contents, Sejong University, Seoul, 05006, South Korea
[2]Department of Software Convergence, Sejong University, Seoul, 05006, South Korea
[3]Department of Artificial Intelligence and Linguistic Engineering, Sejong University, Seoul, 05006, South Korea
[4]Communication & Media Research Laboratory, Electronics and Telecommunications Research Institute, Daejeon, 34129, South Korea
[5] Department of Software, Collage of Software Convergence, Sejong University, Seoul, 05006, South Korea
*msyoon@etri.re.kr
**wikim@sejong.ac.kr

**ABSTRACT**

In this paper, we propose a novel, convolutional neural network model to extract highly precise depth maps from missing viewpoints, especially well applicable to generate holographic 3D contents. The depth map is an essential element for phase extraction which is required for synthesis of computer-generated hologram (CGH). The proposed model called the HDD Net uses *MSE* for the better performance of depth map estimation as loss function, and utilizes the bilinear interpolation in up sampling layer with the Relu as activation function. We design and prepare a total of 8,192 multi-view images, each resolution of 640 by 360 for the deep learning study. The proposed model estimates depth maps through extracting features, up sampling. For quantitative assessment, we compare the estimated depth maps with the ground truths by using the *PSNR*, *ACC,* and *RMSE*. We also compare the CGH patterns made from estimated depth maps with ones made from ground truths. Furthermore, we demonstrate the experimental results to test the quality of estimated depth maps through directly reconstructing holographic 3D image scenes from the CGHs.

## Introduction

A depth map image represents information related to a distance between the camera's viewpoint and the object's surface. It is reconstructed based on the original (RGB color) image and generally has a gray scale format. Depth maps are used in three-dimensional computer graphics, such as three-dimensional image generation and computer-generated holograms (CGHs). In particular, phase information which is an essential element for computer-generated holograms can be acquired through depth maps [1, 2].

In order to observe 360 degree digital holographic content, 360 degree RGB image, depth map image pairs are required. Without a depth map at some location (missing viewpoint), hologram at that location is also not visible. Thus, depth map estimation for missing viewpoints contributes to the realistic of 360 degree digital hologram content. In this study, we propose a novel method to learn & estimate depth information from RGB image where captured & missing viewpoints. The depth map can be generated from not only captured viewpoints but also from some missing viewpoints. The method is illustrated in Fig. 1.

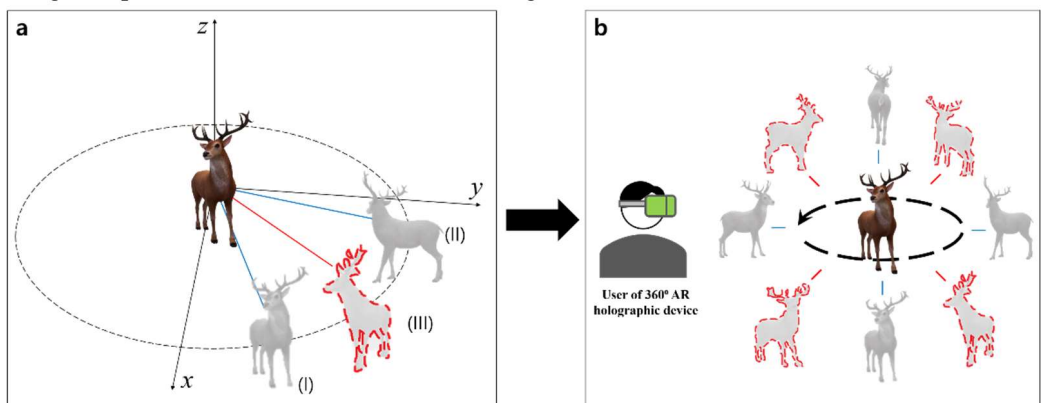

**Figure 1.** **a** Depth map estimation from missing viewpoint (III) using image set where captured viewpoints (I, II), **b** richer holographic 3D contents with estimated depth maps from missing viewpoints (III).

Early studies on estimating depth maps have been introduced. One of them is monocular approach [3-10]. Battiato et al. [3] proposed generating depth maps by image classification. In this work, digital images are classified into indoors / outdoors / outdoors with geometric objects, with low computational costs. Eigen et al. [4] using convolutional neural network (CNN) based model consisting of two different networks. One estimates the global structure of the scene while the other estimates local information. Koch et al. [5] studied preservation of edges and planar regions, depth consistency, and absolute distance accuracy from a single image. Other studies adopting Conditional Random Field [6-8], Generative Adversarial Network [9, 10] and U-nets with Encoder-Decoder structure have been introduced [11]. Alhashim et al. [12] applied transfer learning to high-resolution depth map estimation which we will refer it to Conventional Dense Depth (CDD) for the rest of this paper. Our work to be introduced is based on this abovementioned method.

Another approach is stereo approach where left image and right image are used [13]. In this method, depth maps were estimated based on disparity between two opposite viewpoints (two cameras). This approach follows the intuition of recognizing the distance of an object through human vision perception. Self-supervised learning, where the model estimates merge depth map from stereo images with monocular depth map from left or right image has been applied to this type of approach [14].

Previously, attempts to estimate depth maps using multi-view images (more than 2 RGB images) [15-18] have also been made. There have been studies using the plane-sweep method [19] which is a basic computational geometry algorithm to find intersecting line segments [15-17]. Choi et al. [15] used convolutional neural networks (CNN) for multi-view stereo matching, which combines the cost volumes along the depth hypothesis in multi-view images. Im et al. [16] proposed an end-to-end model which learns a full plane-sweep process including construction of cost volumes. Recently, Zhao et al. [17] proposed an asymmetric encoder-decoder model which has better accuracy for outdoor environments. Wang et al.'s work features a CNN for solving the depth estimation problem on several image-pose pairs taken continuously while camera is moving [18]. We note that the previous works mentioned in this paragraph and the previous one use multi-view color images to generate only a single depth image.

For a user to observe even more realistic holographic content as well as more continuous AR/VR content (see Fig. 1-**b**), it is very critical to estimate as many highly precise depth maps for each of a given narrow angular range as possible in short time. Therefore, in the study, although we use input data of multi-view color images, we do not adopt either conventional stereo-view or multi-view methods since these methods use multiple color images as input in order to output only a single depth map. We instead adopt the monocular approach, which is able to produce a singular depth map estimation from a single color image, utilizing Alhashim's method [12], which is suitable for high-resolution depth map estimation. Thus, we contrive a novel model that learns process of estimating depth maps for given multi-view color images and then estimates new depth maps from new color images of the missing viewpoints as illustrated in (III) of Fig. 1-**a**. Furthermore, we demonstrate the experimental results to test the quality of estimated depth maps through reconstructing 3D image scenes from the CGHs generated by using the estimated depth maps.

## Results

### Data Preparations

We first introduce a method that generate 360 degree, multi-view RGB image-depth map pairs dataset using Z-depth rendering function provided by Maya 2018. Second, we present a neural network architecture that estimates depth maps. Third, experimental results are discussed. In addition, we also show the results of synthesizing CGHs, numerical reconstruction and optical reconstruction using RGB image-depth map pairs in the next section. In this study, a dataset of multi-view's RGB image-depth map was generated using Z-depth rendering provided by Maya software. In order to extract the RGB image-depth map pairs in Maya, we need two identical 3D objects located near the origin and a virtual camera with light source which film these objects, so that the virtual camera could acquire depth difference information between two objects during the camera rotates around the origin. Depth measurements along Z-direction from the camera were made using a Luminance Depth preset of the Maya-supplied. Details of the process are shown in Table 1.

| Distance from virtual camera to 255 depth | 11 cm | Margin from depth spot to object | 2 cm |
|---|---|---|---|
| Distance from virtual camera to 0 depth | 28.7 cm | Distance between two objects (Center of object A to center of object B) | 8.3 cm |
| Distance from 0 depth to 255 depth | 14.2 cm | Radius of camera rotation trajectory | 20 cm |

**Table 1.** Object and camera settings.

The virtual camera acquires 1024 pairs (per each object of both RGB images and depth maps, see Table 2 for details) at every about 0.35 degrees while rotating 360 degree around the y-axis. 3D objects which we used for the study are torus, cube, cone, and sphere as shown Fig. 2.

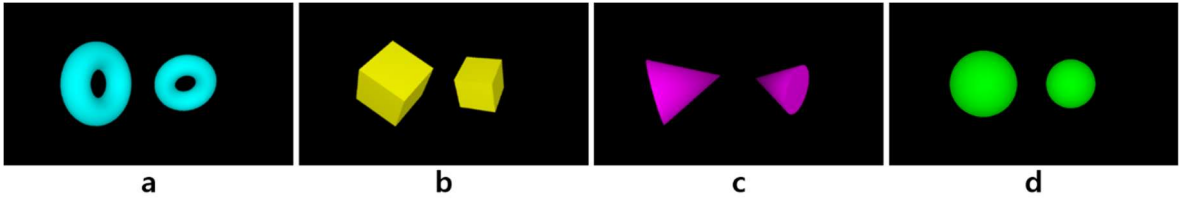

**Figure 2.** Types of 3D objects: **a** Torus, **b** Cube, **c** Cone, **d** Sphere.

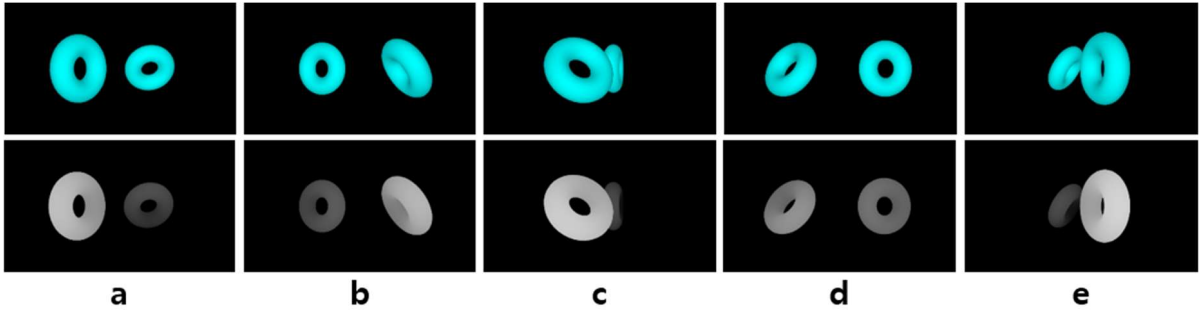

**Figure 3.** Sample images observed at various viewpoints of RGB-Depth ground truth data (Torus): **a** 0 degree, **b** 72 degree, **c** 144 degree, **d** 216 degree, **e** 288 degree.

For the experiments in this study, we used a total of 8,192 images (Table 2). Among them, the images of 4,096 are RGB color images and the remaining images of 4,096 are depth map images. Both color images and depth map images are classified into 4 shapes (torus, cube, cone, and sphere), each consisting of 1,024 views. When 4 kinds of objects were learned at once, 60% interleaved of the total data were used for training, and 40% were used for testing (estimating).

| Generated sample | | Shape | | Training set/Test set | | | |
|---|---|---|---|---|---|---|---|
| Color images | 4,096 | Torus | 1,024 | Set for training (60%) | 614 | Set for test (40%) | 410 |
| | | Cube | 1,024 | | 614 | | 410 |
| | | Cone | 1,024 | | 614 | | 410 |
| | | Sphere | 1,024 | | 614 | | 410 |
| Depth map images | 4,096 | Torus | 1,024 | Set for training (60%) | 614 | Set for test (40%) | 410 |
| | | Cube | 1,024 | | 614 | | 410 |
| | | Cone | 1,024 | | 614 | | 410 |
| | | Sphere | 1,024 | | 614 | | 410 |

**Table 2.** Sample sets in which RGB color images and depth maps are prepared for training and test in the research.

## Quantitative Results

In order to quantitatively evaluate the performance of depth map estimation from the proposed model (HDD: Holographic Dense Depth), first we present a characteristic of the loss *MSE* averaged about entire objects during training for the HDD model in Fig. 4-**a**. Then, in the case of torus, we present the typical trends of *PSNR* (Peak Signal-to-Noise) of HDD and CDD in Fig. 4-**b**, where the *x*-axis represents the step number during 90 epochs. In the case of torus, Fig. 4-**c** shows the distribution of *ACC* (Accuracy) [20], indicating the similarity between ground truth depth map and estimated depth map from HDD or from CDD, where the *x*-axis represents the angular degree corresponding to viewpoint (see Fig. 1). Also, we present *RMSE* from HDD and from CDD for each object (Fig. 4-**d**).

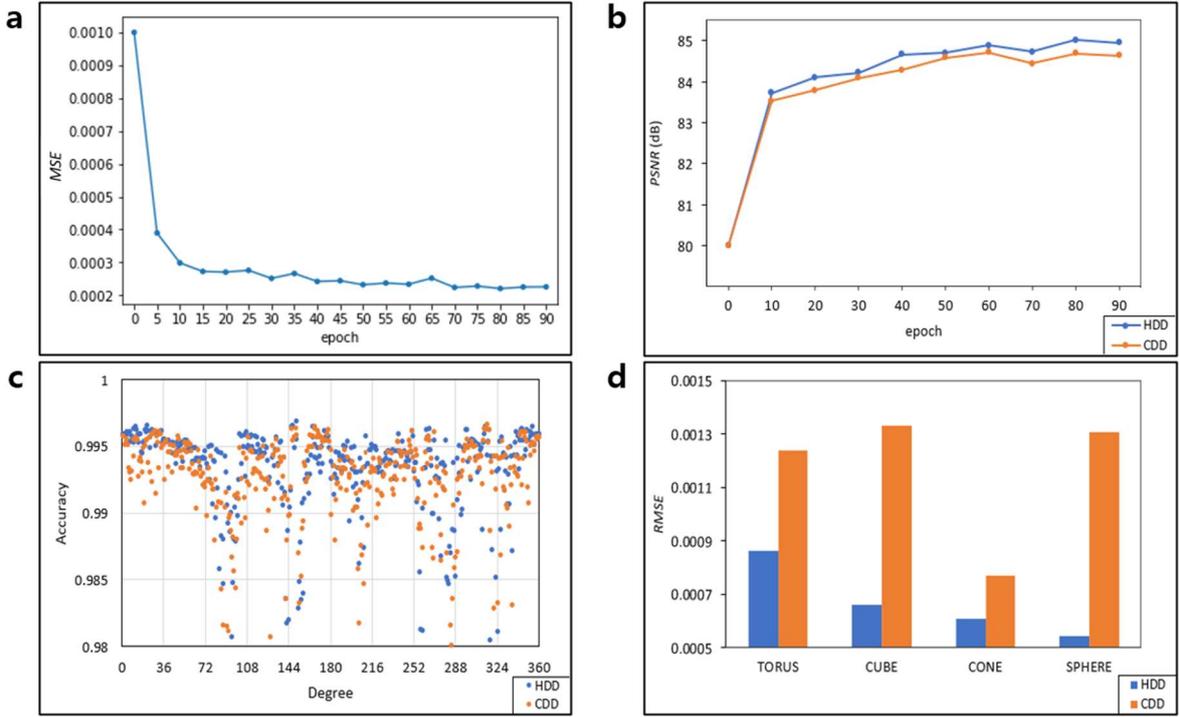

**Figure 4.** **a** The curve of the loss MSE averaged about entire objects during training for the HDD model, **b-d** Comparison of HDD and CDD using *PSNR* trend, *ACC* distribution with ground truth depth map for torus, *RMSE* difference for entire object after training.

We introduce a loss function which we used in study. *MSE* (Mean Squared Error) was used as a loss function of HDD to more efficient gradient transfer. The *MSE* is defined as

$$MSE = \frac{1}{n}\sum_{i=1}^{n}(y_i - y_i')^2$$

(1)

which refers to the value measured by the pointwise operation of the difference between the depth value of estimated depth map ($y_i$) and the depth value of ground truth depth map ($y_i'$), and *n* is a total number of pixels. Quantitative metrics used in the study is as following. The *SSIM* (Structural Similarity) [21] is defined as

$$SSIM = \frac{(2\mu_y\mu_{y'} + c_1)(2\sigma_{yy'} + c_2)}{(2\mu_y^2 + \mu_{y'}^2 + c_1)(\sigma_y^2 + \sigma_{y'}^2 + c_2)}$$

(2)

which is commonly used to evaluate the visual quality (luminance, contrast and structure) similarity with the original image, with $\mu_y$ the average of ground truth(*y*); $\mu_{y'}$ the average of estimated depth map(*y'*); $\sigma_y^2$ the variance of ground truth(*y*); $\sigma_{y'}^2$ the variance of estimated depth map(*y'*); $2\sigma_{yy'}$ the covariance of ground truth(*y*) and estimated depth map(*y'*); $c_1$, $c_2$ two variables to stabilize the division with weak denominator, which is defined as

$$c_1 = (k_1 L)^2, c_2 = (k_2 L)^2$$

(3)

with *L* the dynamic range of the pixel-values (normally $2^{bit\ per\ pixel} - 1$); $k_1$, $k_2$ are defined as default values 0.01 and 0.03 respectively. Both model's *SSIM* is about 0.9999. The *PSNR* is defined as

$$PSNR = 10\ log\frac{s^2}{MSE}$$

(4)

which is defined via the *MSE* and *s* (maximal signal value of given image), equal to 255 for depth map image of 8-bit grey levels which is used in the study. The *ACC*, is defined as

$$ACC = \frac{\sum_d(I \cdot I')}{\sqrt{[\sum_d I^2][\sum_d I'^2]}}$$

(5)

where *I* means depth value of pixels in the depth map image estimated by CDD or HDD, and *I'* means depth value of pixels in the ground truth depth map image. When estimation result and ground truth are identical, $ACC = 1$. However, note that if $I = kI'$ ($k$ is scalar), $ACC = 1$ even if estimation result and ground truth are different. Also, we have additional metrics which were used in prior study [4, 12, 22, 23] as follow. Table 3 shows comparison result for these metrics on HDD and CDD.

$$Abs\ rel\ (\text{Relative Absolute Error}) = \frac{1}{n}\sum_{i=1}^{n}\frac{|y_i - y_i'|}{y_i'} \tag{6}$$

$$Sq\ rel\ (\text{Relative Squared Error}) = \frac{1}{n}\sum_{i=1}^{n}\frac{(y_i - y_i')^2}{y_i'} \tag{7}$$

$$RMSE\ (\text{Root Mean Squared Error}) = \sqrt{\frac{1}{n}\sum_{i=1}^{n}(y_i - y_i')^2} \tag{8}$$

$$LRMSE\ (\text{Log Root Mean Squared Error}) = \sqrt{\frac{1}{n}\sum_{i=1}^{n}(\text{LOG}(y_i) - \text{LOG}(y_i'))^2} \tag{9}$$

where $y_i$ is a depth-value of pixel in estimated depth map, $y_i'$ is a depth-value of pixel in ground truth depth map and n is a total number of pixels.

| Indicators / Models | a SSIM | | | | b PSNR (dB) | | | | c ACC | | | |
|---|---|---|---|---|---|---|---|---|---|---|---|---|
| | Torus | Cube | Cone | Sphere | Torus | Cube | Cone | Sphere | Torus | Cube | Cone | Sphere |
| HDD | 0.9999 | 0.9999 | 0.9999 | 0.9999 | 84.95 | 84.42 | 84.90 | 85.03 | 0.9933 ±0.0040 | 0.9933 ±0.0030 | 0.9928 ±0.0037 | 0.9965 ±0.0012 |
| CDD | 0.9999 | 0.9999 | 0.9999 | 0.9999 | 84.64 | 83.94 | 84.68 | 84.62 | 0.9925 ±0.0039 | 0.9934 ±0.0027 | 0.9926 ±0.0036 | 0.9959 ±0.0013 |

| Indicators / Models | d Abs rel | | | | e Sq rel | | | | f RMSE | | | |
|---|---|---|---|---|---|---|---|---|---|---|---|---|
| | Torus | Cube | Cone | Sphere | Torus | Cube | Cone | Sphere | Torus | Cube | Cone | Sphere |
| HDD | 0.022 | 0.018 | 0.022 | 0.017 | 0.0058 | 0.0046 | 0.0058 | 0.0043 | 0.0009 | 0.0007 | 0.0006 | 0.0005 |
| CDD | 0.019 | 0.017 | 0.016 | 0.017 | 0.0062 | 0.0052 | 0.0061 | 0.0047 | 0.0012 | 0.0013 | 0.0008 | 0.0013 |

| Indicators / Models | g LRMSE | | | | Time | | Resolution |
|---|---|---|---|---|---|---|---|
| | Torus | Cube | Cone | Sphere | Training | Estimating | |
| HDD | 0.0114 | 0.0117 | 0.0111 | 0.0110 | 17 hours | 0.18 seconds | 640 × 360 |
| CDD | 0.0116 | 0.0122 | 0.0113 | 0.0112 | 17 hours | 0.18 seconds | 640 × 360 |

**Table 3.** Quantitative comparison of HDD (proposed model) and CDD (Using Nvidia Titan RTX × 8, **a-c**: higher is better, **d-g**: lower is better)

## Qualitative Results

To qualitatively evaluate the depth map estimation performance of the HDD, we present samples of color images, ground truths, and model's estimation results at viewpoint where not used in training as shown in Fig. 5.

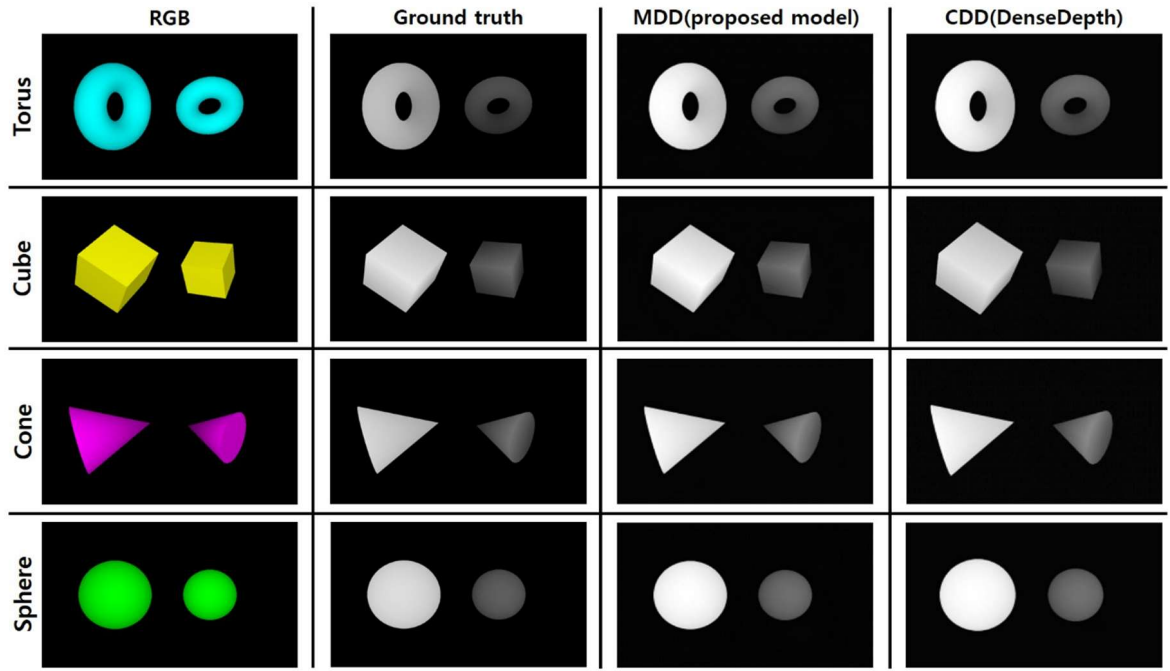

**Figure 5.** Depth map Comparison among ground truth, estimation result from HDD, and estimation result from CDD.

According to ground truth and the model's estimation result, it seems that HDD estimates depth values of the objects quite accurately. However, we can see that the part which closer to camera is slightly brighter than ground truth.

## Hologram image synthesis using depth estimation result

Fig. 6 represent that qualitative comparison on the fringe patterns from depth map estimated by the proposed model and ground truth depth maps.

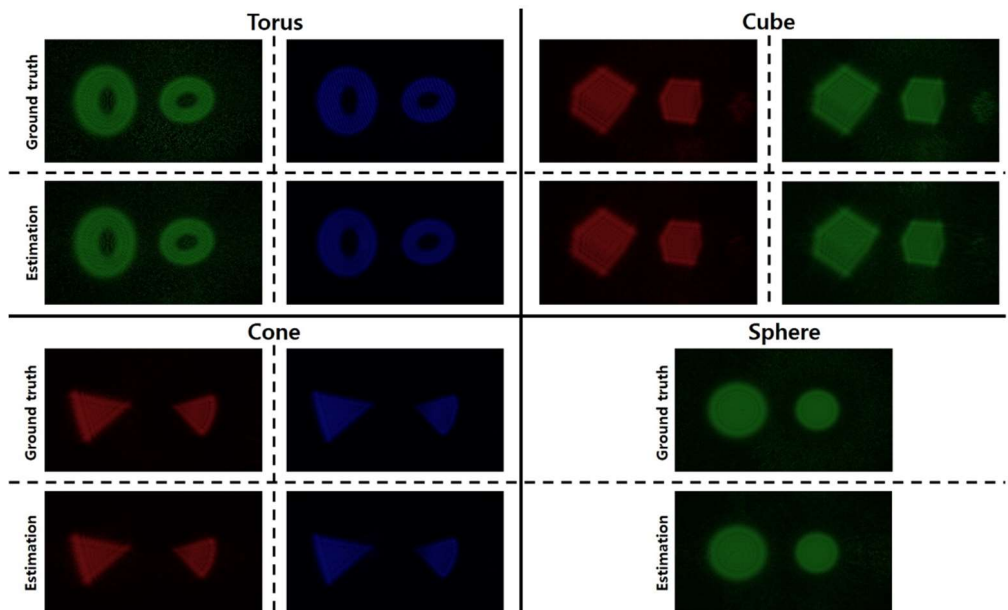

**Figure 6.** Comparison of fringe patterns from estimated depth map and fringe patterns from ground truth depth map for entire object.

Numerically reconstructed hologram images and optically reconstructed hologram images from estimated depth maps and ground truth depth maps are as shown in Fig. 7. (Focused object is indicated by an arrow mark)

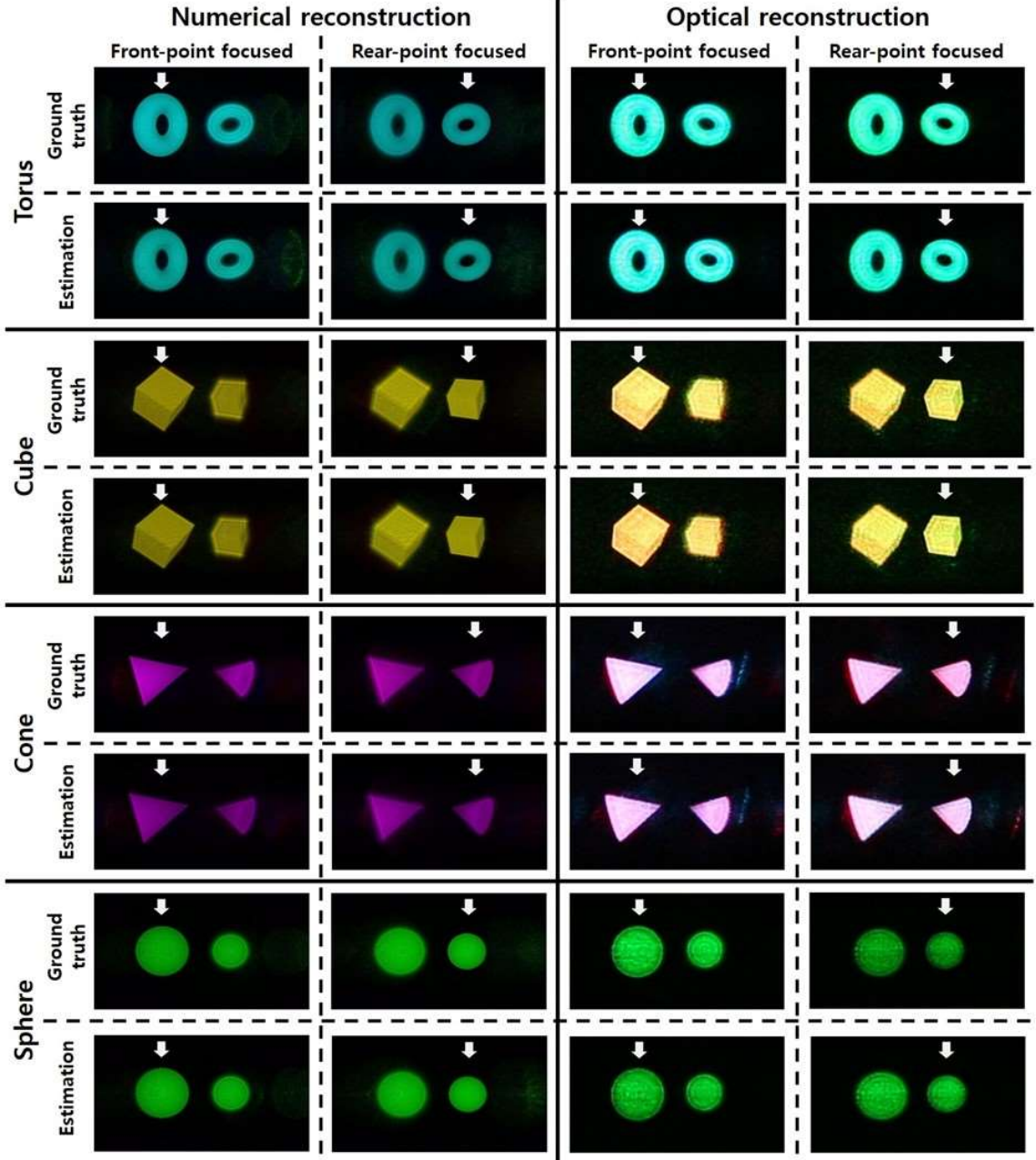

**Figure 7.** Result of Numerical/Optical reconstruction from CGHs using estimated depth map image and ground truth. (Focused object is indicated by an arrow mark)

The CGH was calculated from the RGB & Depth map-based Fast Fourier Transform (FFT) algorithm, with either one set of ground truth depth maps and color images or another set of estimated depth maps and corresponding color images used as input for the computational process of FFT's synthesis [24-25].
CGHs of 1,024 views was prepared for each solid figure, and then a process of encoding called Lee's scheme was adapted so that they can be represented onto an amplitude-modulating typed, spatial light modulator (SLM), that is, a reflective LCoS-SLM [24]. The Lee's encoding is to decompose a complex-valued light field $H(x, y)$ into components with four real and non-negative coefficients, which can be expressed by

$$H(x,y) = L_1(x,y)e^{i0} + L_2(x,y)e^{i\pi/2} + L_3(x,y)e^{i\pi} + L_4(x,y)e^{i3\pi/2}$$
(10)

where at least two among these four coefficients $(L_i)$ are zero. The SLM which used for optical reconstructions,

able to display the grey levels of 8 bits, has the resolution of 3840 x 2160 pixels, the active diagonal length of 0.62", and the pixel pitch of 3.6 μm. A combined beam from RGB laser sources (each wavelength: 638 nm, 520 nm, 450 nm) passes through an expanding & collimating optics to make a coherent, uniform illumination onto the active area of SLM. A field lens (focal length: $f$ = 50 cm) is positioned just after LCoS-SLM; Thus experimental observation of optically reconstructed images was implemented from a DSLR camera (Canon EOD 5D Mark III), whose lens was located within an observation window that was generated near the focus of the field lens [25] (see Fig. 8). The results of camera-captured optical reconstructions as well as numerical reconstructions from synthesized CGHs are illustrated in Fig. 7. In order to prove depth difference in real 3D space between two objects based on these prepared 360 degree holographic contents, we demonstrate the accommodation effect from realized holograms, through presentation to simultaneously indicate the clear object and the blurred object at each picture, when varying camera lens either on rear focal plane or on front focal plane, shown in each photograph in Fig. 8.

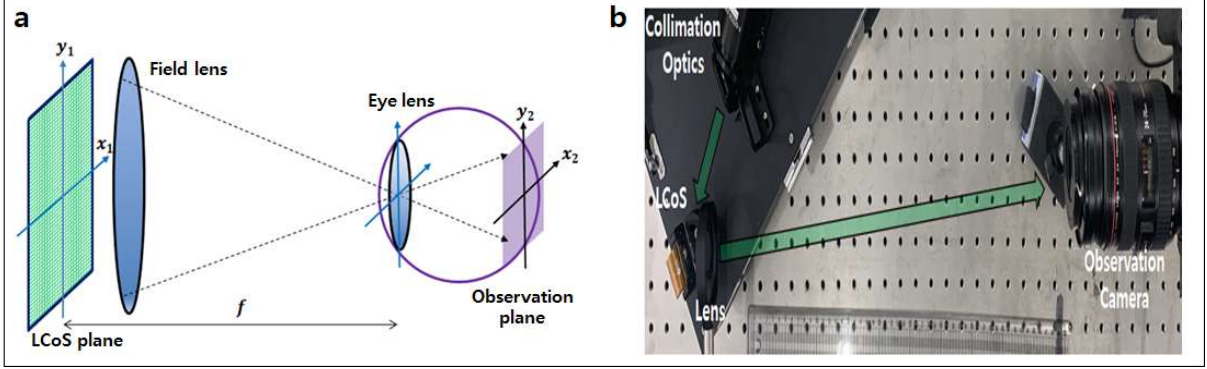

**Figure 8.** Geometry of the optical system for holographic 3D observation: **a** numerical simulation and **b** its optical experiment setup. Observer's eye lens in **a** is located at the position of the focal length of the field lens, corresponding to the center of Fourier plane of the optical holographic display **b**.

When the holographic 3D images prepared from the depth map estimated using the proposed deep learning model are observed, there exists relatively blurring phenomenon, in comparison with those prepared from ground truth depth map. This is because the minute difference between depth values from the two objects on the basis of estimation is not exactly equal to the difference between depth values from the two objects on the basis of reference (ground truth). However, Fig. 7 show that photographs of optically reconstructed scenes clearly support the accommodation effect on holographic 3D images; when an object between two objects is within the camera's focus, its photographic image is very sharp while another object out of focus is completely blurred.

In addition, in order to compare CGH from HDD's estimation depth map with CGH from ground truth depth map quantitatively, we employ the performance evaluation using the *ACC*, which is defined as

$$ACC = \frac{\sum_{r,g,b}(I \cdot I')}{\sqrt{[\sum_{r,g,b} I^2][\sum_{r,g,b} I'^2]}}$$

(11)

where *I* means brightness for each color in CGH image obtained using 2D RGB images and depth map images estimated by HDD, and *I'* means brightness for each color in CGH image obtained using 2D RGB images and ground truth depth map images. When estimation result and ground truth are identical or $I = kI'$ (*k* is positive scalar), then *ACC = 1*. When mismatch between them occurs, $0 \leq ACC < 1$. The comparison results are shown in Fig. 9 and Table 4.

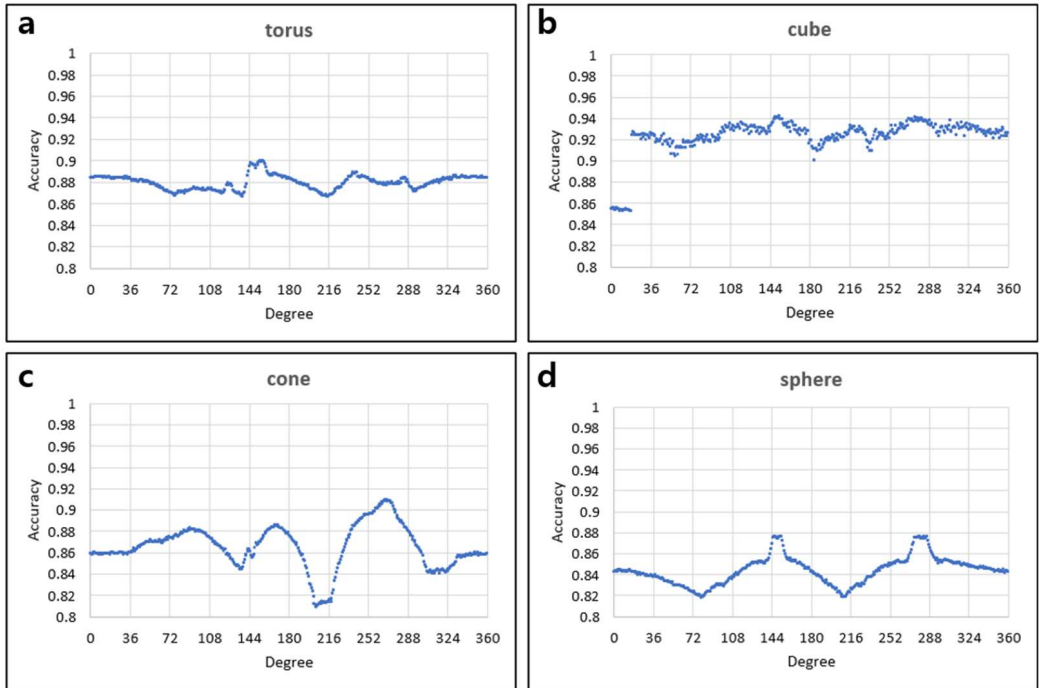

**Figure 9** *ACC*'s trend on CGHs of torus, cube, cone, and sphere.

|     | torus | cube | cone | sphere |
| --- | --- | --- | --- | --- |
| *ACC* | 0.8793 ± 0.0068 | 0.9223 ± 0.0175 | 0.8649 ± 0.0204 | 0.8431 ± 0.0135 |

**Table 4.** *ACC* between CGH from estimated depth map and CGH from ground truth depth map.

## Discussion

In this paper, we proposed and demonstrated a novel, convolutional neural network model that learns estimating depth map from missing viewpoints, especially fit well for holographic 3D. The proposed model, which we call the HDD Net, uses *MSE* for the better performance of depth map estimation in comparison with the CDD that uses *SSIM* 90% and *MSE* 10% as loss function.

We designed and prepared a total of 8,192 multi-view images, each resolution of 640×360 for the purpose. Proposed model, HDD estimates depth maps through extracting features, up sampling. Its weights were optimized by *MSE* loss function. For quantitative assessment, we compared the estimated depth maps from HDD with ones from CDD using the *PSNR*, *ACC*, *RMSE*, *etc*. As seen in Fig. 4, the proposed model HDD is numerically superior to CDD on the metrics of *PSNR*, *ACC*, and *RMSE*. In addition, as shown in Table 3, though HDD is numerically inferior to CDD on the metrics of *abs rel*, superior to CDD on the metrics of *sq rel* and *LRMSE*. We also compared fringe pattern made from estimated depth maps with ones made from ground truths. As seen in Fig. 9 and Table 4, the *ACC* values are about from 0.84 to 0.92 depending on figures, which means that the fringe pattern made from estimated depth map by proposed model is significantly similar to the fringe pattern made from ground truth depth map.

Image resolutions and extraction speed for deep learning-based depth map can be improved by adjusting the model's parameters such as filters and filter's sizes, and then optimizing the ratio of training/test data. Moreover, we remark that only the diffraction efficiency element, *i.e.*, direct observation of the reconstructed holographic 3D image was used as a comparative measure in this paper with the exception of other variables. We are planning to supplement the measure through further analysis taking parameters such as contrast ratio of intensity, clearness and distortion into account to evaluate H3D images.

The contributions of this study are as follows: First, we demonstrate the ability of our proposed HDD to learn and produce depth map estimation of higher accuracy from multi-view color images. Second, we prove the capability to apply deep learning-based, estimated depth maps to synthesize CGHs, with which we can quantitatively evaluate the degree of accuracy in the performance of our proposed model for holography. Third, we illustrate the effectiveness of CGHs synthesized via the proposed HDD by making directly numerical/optical observations of holographic 3D images.

Limitations of the proposed model covered in this study are that there are the minute residual images near the border area of each object and weak background noise in estimated depth maps. In order to overcome these issues, one needs to find approaches to place relatively large weights on these border areas and then to enhance precision estimations of these areas. It is also worth mentioning that image resolutions and extraction speed for deep learning-based depth map can be improved by adjusting the model's parameters such as filters and filter's sizes and then optimizing the ratio of training/test data. Moreover, we remark that only the diffraction efficiency element, *i.e.*, direct observation of the reconstructed holographic 3D image was used as a comparative measure for the quality of H3D image in this study. We are planning to supplement the measure through further analysis taking parameters such as contrast ratio of intensity, clearness and distortion into account to evaluate H3D images.

## Methods

Proposed depth map estimation model HDD consists of two components which are the encoder and the decoder as shown in Fig. 10. We adopt the Dense Depth (CDD) of Alhashim et al. [12], of which the detail architecture is shown in Table 5. The encoder performs feature extraction and down sampling for the input RGB images. The decoder performs up sampling by concatenating the extracted features based on the size of the RGB image. Weights for both components are optimized by loss function which minimizes discrepancy between ground truth and estimated depth map. The CDD learns and estimates depth from a single viewpoint. On the other hand, the HDD learns depth from multiple viewpoints and estimates depth of viewpoints which is not used for train (new viewpoints). The CDD model used *SSIM* 90% and *MSE* 10% as loss function, but the proposed HDD model only used *MSE* as loss function. We also adopted the CDD through utilizing the bilinear interpolation in up sampling layer and the Relu as activation fuction. As a result, we obtained better depth estimation results.

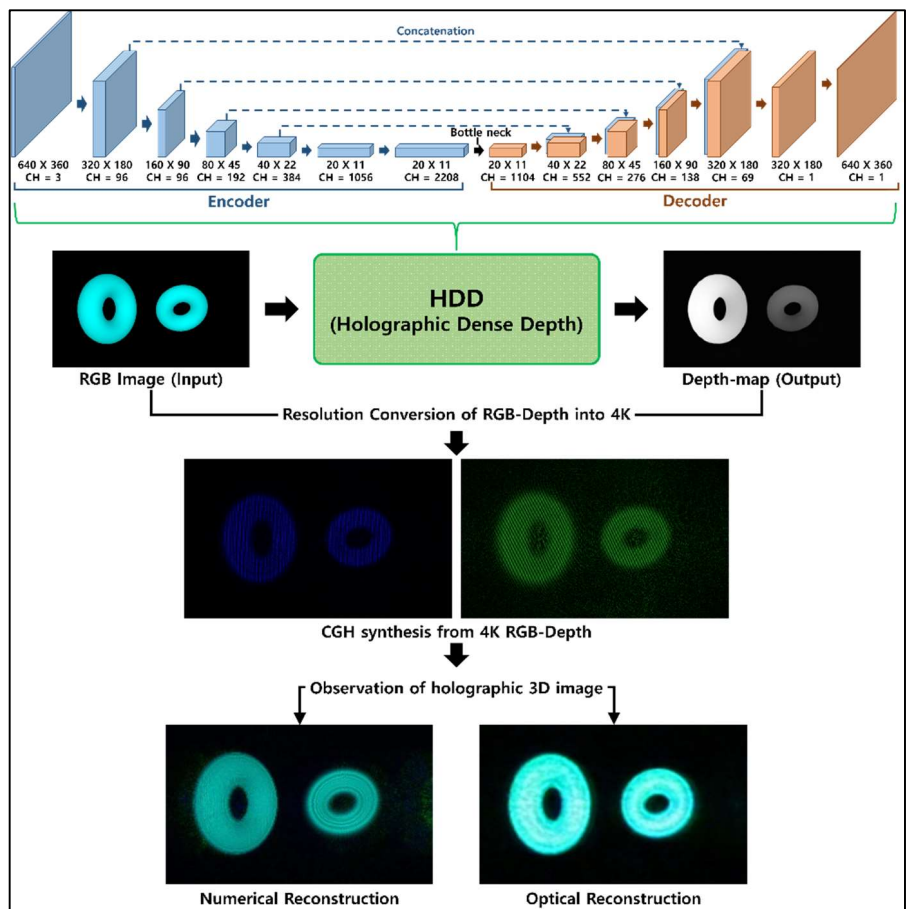

**Figure 10.** Pipeline for Depth map estimation model and digital hologram reconstruction.

| Architecture of proposed model | | |
|---|---|---|
| Input | Input (single RGB image) | [ch = 3, shape = 640×360] |
| Convolution | 7×7 convolution, stride 2 | [ch = 96, shape = 320×180] |
| Encoder (Pre-trained DenseNet-161) | Batch normalization ReLu 3×3 max pooling | [ch = 96, shape = 160×90] |
| | Dense Block (6 Dense layers) Transition layer | [ch = 192, shape = 80×45] |
| | Dense Block (12 Dense layers) Transition layer | [ch = 384, shape = 40×22] |
| | Dense Block (36 Dense layers) Transition layer | [ch = 1056, shape = 20×11] |
| | Dense Block (24 Dense layers) | [ch = 2208, shape = 20×11] |
| | Batch normalization | [ch = 2208, shape = 20×11] |
| Bottleneck | 1×1 convolution | [ch = 1104, shape = 20×11] |
| Decoder (Dense Depth) | Up sampling layer | [ch = 552, shape = 40×22] |
| | Up sampling layer | [ch = 276, shape = 80×45] |
| | Up sampling layer | [ch=138, shape = 160×90] |
| | Up sampling layer | [ch = 69, shape = 320×180] |
| Convolution | 3×3 convolution | [ch = 1, shape = 320×180] |
| Output | Bilinear interpolation | [ch = 1, shape = 640×360] |
| Dense layer | Transition layer | Up sampling layer |
| Batch normalization | Batch normalization | Bilinear interpolation |
| ReLu | ReLu | Skip connection |
| 1×1 convolution | 1×1 convolution | 3×3 convolution |
| Batch normalization | 2×2 max pooling | Batch normalization |
| ReLu | - | ReLu |
| 3×3 convolution | - | 3×3 convolution |
| - | - | Batch normalization |
| - | - | ReLu |

**Table 5.** Main components in the proposed model HDD